\documentclass[letterpaper]{article} 
\usepackage{aaai24}
\usepackage{times}  
\usepackage{helvet}  
\usepackage{courier}  
\usepackage[hyphens]{url}  
\usepackage{graphicx} 
\usepackage{natbib}  
\usepackage{caption} 
\usepackage{algorithm}
\usepackage{algorithmic}

\usepackage{newfloat}
\usepackage{listings}
\DeclareCaptionStyle{ruled}{labelfont=normalfont,labelsep=colon,strut=off} 
\floatstyle{ruled}
\newfloat{listing}{tb}{lst}{}
\floatname{listing}{Listing}

\usepackage{amsmath}
\usepackage{amssymb}
\usepackage{booktabs}
\usepackage{array}
\usepackage{soul}
\usepackage{multirow}
\usepackage{subcaption}
\usepackage{xcolor}

\usepackage{pifont}
\newcommand{\cmark}{\ding{51}}
\newcommand{\xmark}{\ding{55}}

\renewcommand{\paragraph}[1]{
     \noindent{\textbf{#1}} 
 }
 
\usepackage{kotex}

\usepackage[capitalize]{cleveref}
\crefname{section}{Sec.}{Secs.}
\Crefname{section}{Section}{Sections}
\Crefname{table}{Table}{Tables}
\crefname{table}{Tab.}{Tabs.}

\title{Gaussian Mixture Proposals with Pull-Push Learning Scheme \\
to Capture Diverse Events for Weakly Supervised Temporal Video Grounding}
\author {
    Sunoh Kim\textsuperscript{\rm 1},
    Jungchan Cho\textsuperscript{\rm 2},
    Joonsang Yu\textsuperscript{\rm 3},
    YoungJoon Yoo\textsuperscript{\rm 3},
    Jin Young Choi\textsuperscript{\rm 1}
}
\affiliations {
    \textsuperscript{\rm 1}ASRI, Dept. of Electrical and Computer Eng., Seoul National University\\
    \textsuperscript{\rm 2}School of Computing, Gachon University, \\
    \textsuperscript{\rm 3}NAVER CLOVA\\
    \{suno8386, jychoi\}@snu.ac.kr, thinkai@gachon.ac.kr, \{joonsang.yu, youngjoon.yoo\}@navercorp.com
}

\begin{document}

\maketitle

\begin{abstract}
In the weakly supervised temporal video grounding study, previous methods use predetermined single Gaussian proposals which lack the ability to express diverse events described by the sentence query. To enhance the expression ability of a proposal, we propose a Gaussian mixture proposal (GMP) that can depict arbitrary shapes by learning importance, centroid, and range of every Gaussian in the mixture. In learning GMP, each Gaussian is not trained in a feature space but is implemented over a temporal location. Thus the conventional feature-based learning for Gaussian mixture model is not valid for our case. In our special setting, to learn moderately coupled Gaussian mixture capturing diverse events, we newly propose a pull-push learning scheme using pulling and pushing losses, each of which plays an opposite role to the other. The effects of components in our scheme are verified in-depth with extensive ablation studies and the overall scheme achieves state-of-the-art performance.
Our code is available at {https://github.com/sunoh-kim/pps}.
\end{abstract}

\section{Introduction}
Temporal video grounding is a challenging task in computer vision, where the goal is to find the temporal location of starting and ending points described by a sentence query in an untrimmed video.
The task has potential for applications such as video understanding~\cite{carreira2017quo}, video summarization~\cite{ma2002user}, and video retrieval~\cite{dong2019dual}, because it can automatically extract temporal video locations of interest described by given sentences.
For temporal video grounding, a fully supervised approach has made remarkable progress~\cite{kim2021plrn,kim2022swag,gao2017tall}
but require manual annotations of temporal locations for every video-sentence pair.
These manual annotations are usually labor-intensive and noisy due to the subjectivity of annotators, which limits their scalability to real-world scenarios and makes trained models biased~\cite{yuan2021closer, zhou2021embracing}.

\begin{figure}[t!]
  \centering
  \includegraphics[width=\linewidth]{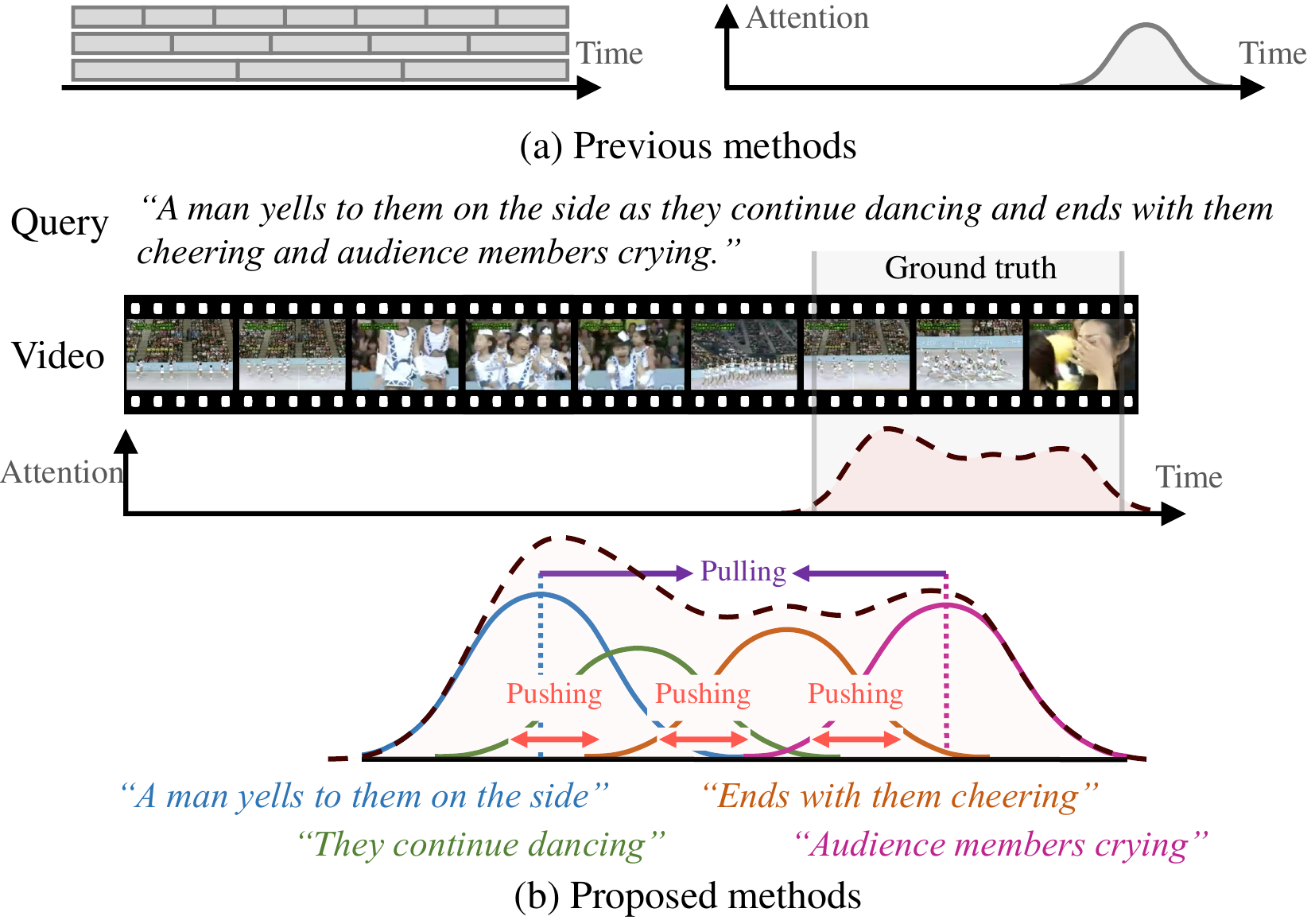}
  \caption{Weakly supervised temporal video grounding. (a) Previous methods use sliding windows (left) or a single Gaussian proposal (right), which has a predetermined shape. (b) The proposed method generates a Gaussian mixture proposal trained to be moderately coupled with a pull-push learning scheme to capture diverse query-relevant events.
  }
\label{fig:concept-art}
\end{figure}

To overcome the limitation, a weakly supervised approach has been proposed to solve the temporal video grounding problem, where only video-sentence pairs are required for training. 
Some existing methods~\cite{huang2021cross, lin2020weakly, mithun2019weakly, tan2021logan, wang2021weakly, zhang2020counterfactual} use a sliding window strategy to generate proposals for a temporal location but use a lot of pre-defined proposals, which require heavy computation.
To reduce the required number of proposals, \cite{zheng2022cnm, zheng2022cpl} generate learnable Gaussian proposals. 
However, these single Gaussian proposals with a peak at its center lack the expression ability for diverse query-relevant events in a video.

To enhance the expression ability, we propose a Gaussian mixture proposal (GMP) that can depict arbitrary shapes by learning importance, centroid, and range of every Gaussian in the mixture.
Since our GMP is implemented over a temporal location, conventional feature-based learning for Gaussian mixture model~\cite{zong2018deep, lee2018simple} is not applicable to our approach.
In our special setting, our goal is to train the GMP to capture a temporal location semantically 
relevant to a sentence query that includes diverse events coupled moderately. 
In \cref{fig:concept-art}, for instance, one sentence query includes two semantic events coupled by ``A man yells to them on the side" and ``They continue dancing".

To capture the 
coupled events in a query, we propose a Pull-Push Scheme (PPS) to learn a GMP whose Gaussians are moderately coupled.
Specifically, we first define a GMP with learnable parameters: importance, centroid, and range of every Gaussian in the mixture.
To learn the importance, we propose an importance weighting strategy that represents importance levels of each Gaussian mask for a query-relevant location.
To generate the GMP that represents a query-relevant location, our PPS is trained to reconstruct the sentence query from the proposal.
In our scheme, the Gaussians in one GMP should be located near a query-relevant temporal location, 
but should not be overlapped too much with others to represent diverse events.
To this end, our scheme leverages a pulling loss and a pushing loss, each of which plays an opposite role to the other to produce moderately coupled Gaussians.
The pulling loss lets the Gaussians stay close to each other by pulling the Gaussian centroids together.
The pushing loss prevents the Gaussians from overlapping too much with the others by forcing the Gaussians to be less overlapped.

We verify that our scheme generates high-quality proposals that significantly improve recall rates on the Charades-STA~\cite{gao2017tall} and ActivityNet Captions~\cite{krishna2017dense} datasets.
We also demonstrate the effectiveness of each component in our scheme with extensive ablation studies.
In summary, our contributions are as follows.
\begin{itemize}
\item We generate a Gaussian mixture proposal that represents a query-relevant temporal location by learning importance, centroid, and range of every Gaussian to enhance the expression ability of the proposal.
\item We propose a pull-push learning scheme that uses a pulling loss and a pushing loss, each of which plays an opposite role to the other to capture diverse events.
\item The proposed components are verified in-depth with extensive ablation studies and the overall scheme achieves state-of-the-art performance. 
\end{itemize}

\section{Related Work}
\begin{figure*}[t!]
  \centering
  \includegraphics[width=\linewidth]{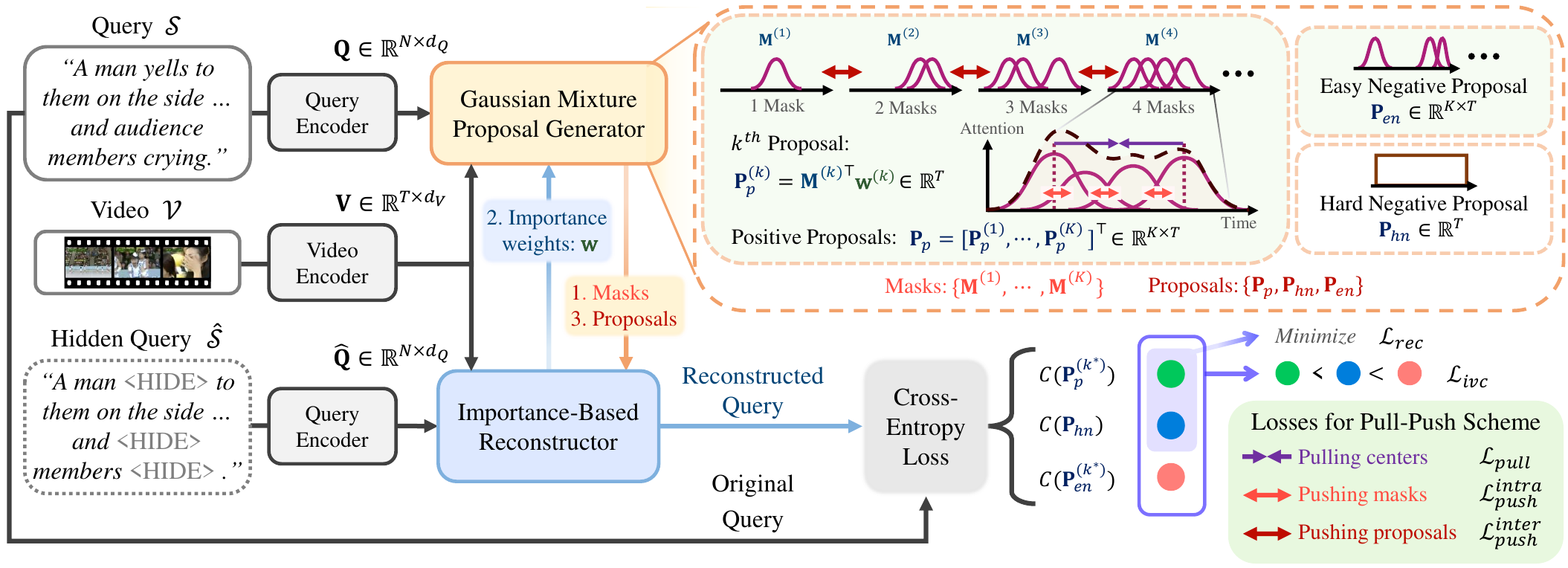}
  \caption{
  The overall scheme of the proposed method.
  The Gaussian mixture proposal generator produces multiple Gaussian masks from the features representing both the video and sentence query.
  For the positive proposals, we define a Gaussian mixture proposal, where multiple Gaussian masks are combined via attentive pooling using the importance weights from the importance-based reconstructor.
  Further, to generate moderately coupled masks in the mixture proposal,
  we propose the pull-push learning scheme using $\mathcal{L}_{pull}$, $\mathcal{L}^{intra}_{push}$, and $\mathcal{L}^{inter}_{push}$.
  The importance-based reconstructor leverages the proposals to produce the reconstructed query from the hidden query. 
  }
\label{fig:framework}
\end{figure*}

\subsection{Weakly Supervised Temporal Video Grounding}
\label{sec:weakly-supervised-temporal-video-grounding}


\paragraph{Sliding window-based methods}~\cite{huang2021cross, tan2021logan, mithun2019weakly, wang2021weakly, zhang2020counterfactual} generate proposals through the sliding window strategy and select the most probable proposal.
\cite{tan2021logan} proposes a multi-level co-attention model to learn visual-semantic representations.
\cite{huang2021cross} uses relations between sentences to understand cross-moment relations in videos.
However, sliding window-based methods make a lot of proposals with a predefined length and use Non-Maximum Suppression (NMS)~\cite{neubeck2006efficient} to reduce redundant proposals.
This process requires a large amount of computation. 
The proposed method generates learnable Gaussian mixture proposals without using the sliding window.

\paragraph{Reconstruction-based methods}~\cite{lin2020weakly, zheng2022cnm, zheng2022cpl, song2020weakly, cao2023iterative} assume that well-generated proposals can reconstruct a sentence query from a randomly hidden sentence query.
Early works~\cite{lin2020weakly, song2020weakly} aggregate contextual information of video-sentence pairs to score proposals sampled at different scales.
However, these methods need to select one proposal from a large set of proposals, which requires heavy computation costs.
To solve this problem, other reconstruction-based methods~\cite{zheng2022cnm, zheng2022cpl} propose a learnable Gaussian proposal for a small set of proposals.
\cite{cao2023iterative} iteratively refines proposal confidence scores to prevent the grounding results from being biased.
Unlike the previous methods, our goal is to enhance the expression ability of proposals, hence we generate Gaussian mixture proposals which can effectively represent an arbitrary shape.

\subsection{Gaussian-based Approach}
\label{sec:gaussian-proposals-in-video}

Gaussians have been studied in various tasks~\cite{zong2018deep, lee2018simple, piergiovanni2019temporal, long2019gaussian, zheng2022cnm, zheng2022cpl}.
For weakly-supervised temporal video grounding, \cite{zheng2022cnm, zheng2022cpl} propose learnable Gaussian proposals.
Specifically, \cite{zheng2022cnm} generates one Gaussian proposal for one temporal location, and \cite{zheng2022cpl} generates multiple Gaussian proposals and selects one proposal to predict a query-relevant temporal location.
For action localization, 
\cite{long2019gaussian} uses multiple Gaussian proposals to localize multiple actions, where each single Gaussian proposal represents a temporal location of a specific action.

However, a single Gaussian is a pre-determined shape with a high value at its center, which is not suitable for expressing diverse query-relevant events.
To effectively represent the diverse events, we propose a Gaussian mixture proposal by learning importance, centroid, and range of every Gaussian in the mixture.
For action localization,
\cite{piergiovanni2019temporal} proposes a layer of Gaussian mixture that replaces a conventional convolutional layer to extract video features.
Also, there have been various tasks that train Gaussian mixture model in a feature space~\cite{zong2018deep, lee2018simple}.
Unlike these methods, we generate Gaussian mixture proposals that are directly implemented over a temporal location.
To represent a query-relevant temporal location that has diverse events, we propose a pull-push scheme to learn the moderately coupled Gaussian mixture.

\section{Proposed Method}

The overall scheme of the proposed method is depicted  in~\cref{fig:framework}.
We generate a new proposal model using multiple learnable Gaussian masks from a video feature $\mathbf{V}$ and a query feature $\mathbf{Q}$.
Here, each mask in a video plays a role in focusing on a specific video event and suppressing the rest.
We use a mixture model consisting of multiple Gaussian masks to produce proposals.
Each positive proposal is called {\it Gaussian mixture proposal} (GMP). 
To generate $K$ GMPs ($\mathbf{P}_{p}$), we propose an importance weighting strategy to represent importance levels of each Gaussian mask for a query-relevant location.
For the importance weighting strategy, the importance-based reconstructor receives the generated Gaussian masks and estimates the importance weights of the Gaussian masks for the mixture.
Then, the GMP is obtained via attentive pooling with the Gaussian masks and importance weights.
To capture diverse query-relevant events, we propose a pull-push learning scheme, where the Gaussian masks are trained by pulling loss and pushing loss. 
The pulling loss $\mathcal{L}_{pull}$ makes the masks in a GMP be densely overlapped, whereas  
the pushing loss $\mathcal{L}_{push}$ makes the masks in a GMP be less overlapped.
Each of $K$ easy negative proposals ($\mathbf{P}_{en}$) is also composed of multiple Gaussian masks to capture diversely-shaped confusing locations within the given video. 
Unlike the positive proposal, the easy negative proposal does not use importance weights because the importance weights only represent query-relevant levels, which is only needed for positive proposals.
The importance-based reconstructor receives positive proposals from the Gaussian mixture proposal generator and reconstructs the sentence query from a randomly hidden sentence query.

\subsection{Encoders}
\label{sec:encoders}

Given a video and sentence query, we use pre-trained encoders to obtain a video feature and a query feature, following previous methods~\cite{wu2020reinforcement,chen2021towards}

\paragraph{Video encoder.}
An untrimmed raw video $\mathcal{V}$ is made into a video feature $\mathbf{V}$ through the pre-trained 3D Convolutional Neural Network (3D CNN)~\cite{carreira2017quo, tran2015learning}.
The video feature $\mathbf{V}$ is given by
$\mathbf{V}=[\mathbf{v}_1,\mathbf{v}_2,\dots,\mathbf{v}_T]^\top \in \mathbb{R}^{T\times d_V} \text{,}$
where $\mathbf{v}_t$ is the $t^{th}$ segment feature, $T$ is the number of video segments, and $d_V$ is the dimension of the segment feature $\mathbf{v}_t$.

\paragraph{Query encoder.}
Given a sentence query $\mathcal{S}$, we use the pre-trained GloVe~\cite{pennington2014glove} word embedding to obtain a query feature $\mathbf{Q}=[\mathbf{q}_1,\mathbf{q}_2,\dots,\mathbf{q}_N]^\top \in \mathbb{R}^{N\times d_Q} \text{,}$
where $\mathbf{q}_n$ is the $n^{th}$ word feature, $N$ is the number of words, and $d_Q$ is the dimension of the word feature $\mathbf{q}_n$.

\subsection{Gaussian Mixture Proposal Generator}
\label{sec:proposal-generator}

From video and query features $\mathbf{V}$ and $\mathbf{Q}$, the proposed  GMP  generator yields $K$ positive GMPs ($\mathbf{P}_{p}$), $K$ easy negative proposals ($\mathbf{P}_{en}$) in addition to one existing  hard negative proposal ($\mathbf{P}_{hn}$).

\paragraph{Modeling of GMP for positive proposal.}
For the generation of the positive proposal, we first extract a multi-modal feature $\mathbf{G}$ reflecting both visual and textual information.
We use a transformer~\cite{vaswani2017attention} to aggregate the information of $\mathbf{V}$ and $\mathbf{Q}$ by
\begin{math}
  \mathbf{G}=f_{td}(\widehat{\mathbf{V}}, f_{te}(\mathbf{Q})) =[\mathbf{g}_1,\mathbf{g}_2,\dots,\mathbf{g}_{T},\mathbf{g}_{cls}]^\top\in \mathbb{R}^{(T+1)\times d_G}\text{,}
\end{math}
where the transformer uses $\mathbf{Q}$ as an input to the transformer encoder $f_{te}(\cdot)$ and both $\widehat{\mathbf{V}}$ and $f_{te}(\mathbf{Q})$ as inputs to the transformer decoder $f_{td}(\cdot)$, and $d_G$ is the dimension of the multi-modal feature.
For the video feature $\widehat{\mathbf{V}}$, we append a learnable token $\mathbf{v}_{cls}$, same as a [CLASS] token in \cite{devlin2018bert}, by
$\widehat{\mathbf{V}}=[\mathbf{v}_1,\mathbf{v}_2,\dots,\mathbf{v}_T, \mathbf{v}_{cls}]^\top \in \mathbb{R}^{(T+1)\times d_V}$.
By the transformer, correspondingly, the vector $\mathbf{g}_{cls}\in \mathbb{R}^{d_G}$ stores the sequence information of all words and video segments.

For the $k^{th}$ positive proposal $\mathbf{P}_{p}^{(k)}$, we define multiple Gaussian masks $\mathbf{M}^{(k)} = [\mathbf{M}_{1}^{(k)},\mathbf{M}_{2}^{(k)},\dots,\mathbf{M}_{E_p}^{(k)}]^\top \in \mathbb{R}^{E_p\times T}$, where $\mathbf{M}_{l}^{(k)}$ is the $l^{th}$ Gaussian mask for $\mathbf{M}^{(k)}$, and $E_p$ is the number of masks.
The $k^{th}$ proposal $\mathbf{P}_{p}^{(k)}$ is defined by a mixture of the Gaussian masks $\mathbf{M}^{(k)}$.
The Gaussian centers $\mathbf{c}^{(k)}$ and widths (standard deviations) $\mathbf{s}^{(k)}$ of $\mathbf{M}^{(k)}$ are calculated by the function of $\mathbf{g}_{cls}$, as
\begin{align}
  &\mathbf{c}^{(k)} = \textrm{Sigmoid}\left(\mathbf{W}_\mathbf{c}\,\mathbf{g}_{cls}+\mathbf{b}_\mathbf{c}\right) \in \mathbb{R}^{E_p} \text{,} 
  \label{eq:mask-center}\\
  &\mathbf{s}^{(k)} = \frac{1}{\sigma}\textrm{Sigmoid}\left(\mathbf{W}_\mathbf{s}\,\mathbf{g}_{cls}+\mathbf{b}_\mathbf{s}\right) \in \mathbb{R}^{E_p} \text{.}
  \label{eq:mask-width}
\end{align}
Here, $\mathbf{W}_\mathbf{c~or~s}$ and $\mathbf{b}_\mathbf{c~or~s}$ are defined as  learnable parameters of a fully connected layer, and $\sigma$ is a hyper-parameter controlling the width of the masks.
Consequently, we obtain the $l^{th}$ Gaussian mask $\mathbf{M}_{l}^{(k)} = [f_{l}^{(k)}(0), f_{l}^{(k)}(1), \dots, f_{l}^{(k)}(T-1)]^\top \in \mathbb{R}^{T}$ using 
\begin{equation}
  f_{l}^{(k)}(t) = \mathrm{exp}\left(-\left(\frac{t/(T-1)-\mathbf{c}^{(k)}_{l}}{\mathbf{s}^{(k)}_{l}}\right)^2\right) \text{,} 
  \label{eq:gaussian}
\end{equation}
where $\mathbf{c}^{(k)}_{l}$, $\mathbf{s}^{(k)}_{l} \in \mathbb{R}$ are the $l^{th}$  elements of $\mathbf{c}^{(k)}$, $\mathbf{s}^{(k)}$, respectively.

The $k^{th}$ proposal $\mathbf{P}_{p}^{(k)}$ is defined by a mixture of the Gaussian masks $\mathbf{M}^{(k)}$ via attentive pooling with mask importance weights $\mathbf{w}^{(k)}\in\mathbb{R}^{E_p}$. 
Finally, we generate $K$ positive proposals $\mathbf{P}_{p} = [ \mathbf{P}_{p}^{(1)}, \mathbf{P}_{p}^{(2)}, \dots, \mathbf{P}_{p}^{(K)} ]^\top \in \mathbb{R}^{K\times T}$, where the $k^{th}$ proposal is
\begin{equation}
  \mathbf{P}_{p}^{(k)} = \mathbf{M}^{(k)\top}\mathbf{w}^{(k)} \in \mathbb{R}^{T} \text{.}
  \label{eq:gaussian-mixture-proposal}
\end{equation}
To represent the importance levels of each Gaussian mask in the mixture, we leverage an importance weighting strategy, where the importance weights $\mathbf{w}^{(k)}$ are estimated by the importance-based reconstructor in \cref{eq:mask-importance-weight}.

\paragraph{Losses for pull-push learning scheme.}
In our scheme, the Gaussian masks in a Gaussian mixture proposal should be densely located near a query-relevant temporal location, but should not be overlapped too much with each other to represent diverse events.
To this end, we propose a pull-push learning scheme using a pulling loss and a pushing loss, each of which plays an opposite role to the other, to produce moderately coupled masks.

The pulling loss $\mathcal{L}_{pull}$ lets the masks stay close, which is computed by minimizing the Euclidean distance between the centers of the two farthest masks as follows:
\begin{equation}
  \mathcal{L}_{pull} = \sum_{k=1}^K \left(\mathbf{c}^{(k)}_{l_{min}} - \mathbf{c}^{(k)}_{l_{max}}\right)^2 \text{,}
  \label{eq:pulling-loss}
\end{equation}
where $l_{min}=\mathrm{arg\,min}_l\, \mathbf{c}^{(k)}_{l}$ and $l_{max}=\mathrm{arg\,max}_l\, \mathbf{c}^{(k)}_{l}$.

The pushing loss is defined by two losses: (1) an intra-pushing loss and (2) an inter-pushing loss.
The intra-pushing loss $\mathcal{L}^{intra}_{push}$ prevents the masks in a proposal from overlapping too much with others by forcing the masks to be less overlapped, which ensures each mask represents different events.
Furthermore, we use the inter-pushing loss $\mathcal{L}^{inter}_{push}$ to let each proposal predict different temporal locations. Based on the regularization term in \cite{lin2017structured}, the resultant two pushing losses are given as
\begin{align}
  &\mathcal{L}^{intra}_{push} = \sum_{k=1}^K || \mathbf{M}^{(k)} \mathbf{M}^{(k)\top} - \lambda_1 I||^2_{F} \text{,} 
  \label{eq:intra-pushing-loss} \\
  &\mathcal{L}^{inter}_{push} = || \mathbf{P}_{p} \mathbf{P}_{p}^\top - \lambda_2 I||^2_{F} \text{,}
  \label{eq:inter-pushing-loss}
\end{align}
where $||\cdot||_F$ denotes the Frobenius norm, $I$ is an identity matrix, and $\lambda_1$ and $\lambda_2$ are hyper-parameters controlling the strength of the pushing.

\paragraph{Negative proposal mining.}
To capture diverse shapes of confusing temporal locations inside the video, we generate a new type of a negative proposal with multiple Gaussian masks, called easy negative proposals
($\mathbf{P}_{en} \in \mathbb{R}^{K\times T}$) in addition to the existing hard negative proposal
($\mathbf{P}_{hn} \in \mathbb{R}^{T}$).
To generate $K$ easy negative proposals, we leverage multiple Gaussian masks to include confusing locations.
In our negative proposal mining, the $k^{th}$ easy negative proposal ($\mathbf{P}^{(k)}_{en}$) is composed of multiple Gaussian masks by using the same process in \cref{eq:mask-center,eq:mask-width,eq:gaussian}.
Contrary to moderately coupled Gaussian masks in the positive proposal, we let the $E_{en}$ Gaussian masks of each easy negative proposal spread sparsely without the pull-push learning scheme because most of the confusing locations exist throughout the entire video.
Then, following \cite{zheng2022cnm,zheng2022cpl}, the hard negative proposal $\mathbf{P}_{hn}$ is determined by a mask covering an entire video, which is  $\mathbf{P}_{hn} = [1, 1, \dots, 1] \in \mathbb{R}^{T}$, where both the query-relevant location and confusing locations are included.
Finally, the Gaussian mixture proposal generator produces three proposals $\{\mathbf{P}_{p}, \mathbf{P}_{hn}, \mathbf{P}_{en}\}$.

\subsection{Importance-based Reconstructor}
\label{sec:importance-based-reconstructor}

We propose an importance weighting strategy to effectively represent importance levels of each Gaussian mask in the mixture.
The importance-based reconstructor produces mask importance weights ($\mathbf{w}$) for Gaussian mixture proposals in \cref{eq:gaussian-mixture-proposal}.
Moreover, the reconstructor receives proposals from the generator and reconstructs the sentence query. 

\paragraph{Mask importance.}
We estimate the $k^{th}$ mask importance weights ($\mathbf{w}^{(k)}$) from the Gaussian masks $\mathbf{M}^{(k)}$. 
First, we use a Mask-Conditioned transformer (MC transformer)~\cite{zheng2022cnm,lin2020weakly} to extract the multi-modal feature $\mathbf{R}^\mathbf{M}$ for any video mask $\mathbf{M}$, given the video feature $\mathbf{V}$ and a randomly hidden sentence query feature $\widehat{\mathbf{Q}}$.
In the MC transformer, the mask $\mathbf{M}$ is multiplied by the self-attention map in every self-attention process to focus on the video feature inside the mask.
Additionally, we append a learnable token $\widehat{\mathbf{q}}_{cls}$
, same as a [CLASS] token in \cite{devlin2018bert}, 
to the hidden sentence query feature by
$\widehat{\mathbf{Q}}=[\widehat{\mathbf{q}}_1,\widehat{\mathbf{q}}_2,\dots,\widehat{\mathbf{q}}_N, \widehat{\mathbf{q}}_{cls}]^\top \in \mathbb{R}^{(N+1)\times d_Q} \text{.}$
The resultant multi-modal
feature $\mathbf{R}^\mathbf{M}$ can be calculated as follows:
\begin{equation}
  \mathbf{R}^\mathbf{M}=f_{md}(\widehat{\mathbf{Q}}, f_{me}(\mathbf{V},\mathbf{M}), \mathbf{M}) \in \mathbb{R}^{(N+1)\times d_R} \text{.}
  \label{eq:reconstructive-transformer}
\end{equation}
Here, the MC transformer uses $\mathbf{V}$ and $\mathbf{M}$ as inputs to the transformer encoder ($f_{me}(\cdot)$).
Then, the transformer decoder ($f_{md}(\cdot)$) receives  $\widehat{\mathbf{Q}}$, $f_{me}(\mathbf{V},\mathbf{M})$, and $\mathbf{M}$. 
The dimension of the multi-modal feature is denoted by $d_R$.
In $\mathbf{R}^\mathbf{M}=[\mathbf{r}^\mathbf{M}_1,\mathbf{r}^\mathbf{M}_2,\dots,\mathbf{r}^\mathbf{M}_{N}, \mathbf{r}^\mathbf{M}_{cls}]^\top$, the vector $\mathbf{r}^\mathbf{M}_{cls}$ reflects all words and video segments conditioned by the mask $\mathbf{M}$.
To compute the $k^{th}$ mask importance weights $\mathbf{w}^{(k)}$ in \cref{eq:gaussian-mixture-proposal}, we calculate $\mathbf{r}^{\mathbf{M}_{l}^{(k)}}_{cls}$ using $\mathbf{M}_{l}^{(k)}$ via \cref{eq:reconstructive-transformer} and apply it to a Multi-Layer Perceptron~(MLP) with two layers as follows:
\begin{align}
  &h^{(k)}_l = \mathrm{MLP}\Bigl(\mathbf{r}^{\mathbf{M}_{l}^{(k)}}_{cls}\,\Bigr)\in \mathbb{R} \text{,}
  \label{eq:mask-importance-value} \\
  &\mathbf{w}^{(k)} = \mathrm{Softmax}\Bigl([h^{(k)}_1,h^{(k)}_2,\dots,h^{(k)}_{E_p}]^\top\Bigr) \in\mathbb{R}^{E_p} \text{.}
  \label{eq:mask-importance-weight}
\end{align}

\paragraph{Losses for reconstruction.}
Based on the supposition that properly generated proposals can reconstruct the given sentence query as in \cite{lin2020weakly, song2020weakly}, we reconstruct the sentence query from a randomly hidden sentence query.
First, we generate the multi-modal features $\mathbf{R}^\mathbf{P}$ using the proposed proposals $\mathbf{P}\in\{\mathbf{P}_{p}, \mathbf{P}_{hn}, \mathbf{P}_{en}\}$ by replacing $\mathbf{M}$ with $\mathbf{P}$ in \cref{eq:reconstructive-transformer}.
Then, the reconstructed query is produced using $\mathbf{R}^\mathbf{P}$, and the cross-entropy loss $C(\cdot)$ is used to measure the difference between the reconstructed query and the original query.
Then, we can calculate $C(\mathbf{P}_{p}^{(k)})$, $C(\mathbf{P}_{hn})$, and
$C(\mathbf{P}_{en}^{(k)})$.
For learning to reconstruct the sentence query, following \cite{lin2020weakly}, we use a reconstruction loss which is the cross-entropy losses of the positive proposals and hard negative proposal, where a query-relevant temporal location exists, as
\begin{math}
\mathcal{L}_{rec}=C(\mathbf{P}_{p}^{(k^*)})+C\left(\mathbf{P}_{hn}\right) \text{,}
\end{math}
where
\begin{math}
  k^* = \mathrm{arg\,min}_k\,C(\mathbf{P}_{p}^{(k)}) \text{.}
\end{math}
Furthermore, following \cite{zheng2022cnm}, we perform contrastive learning to distinguish the query-relevant location from the confusing locations captured by the easy negative proposals and the hard negative proposal.
Based on the triplet loss~\cite{wang2014learning}, the intra-video contrastive loss $\mathcal{L}_{ivc}$ is defined as
\begin{math}
  \mathcal{L}_{ivc}=
  \mathrm{max}\bigl(C(\mathbf{P}_{p}^{(k^*)})-C(\mathbf{P}_{hn})+\beta_1,0\bigr)+
  \mathrm{max}\bigl(C(\mathbf{P}_{p}^{(k^*)})-C(\mathbf{P}_{en}^{(k^*)})+\beta_2,0\bigr)
  \text{,}
\end{math}
where $\beta_1$ and $\beta_2$ are hyper-parameters for margins and $\beta_1 < \beta_2$.

\subsection{Training and Inference}
\label{sec:training-and-inference}

\paragraph{Training.}
In an end-to-end manner, we train our network with five loss terms: 1)  reconstruction loss $\mathcal{L}_{rec}$, 2)  intra-video contrastive loss $\mathcal{L}_{ivc}$, 3)  pulling loss $\mathcal{L}_{pull}$, and two pushing losses of 4) intra-pushing loss $\mathcal{L}^{intra}_{push}$ and 5) inter-pushing loss $\mathcal{L}^{inter}_{push}$. Then the total loss is given by
\begin{math}
  \mathcal{L}_{total}=\mathcal{L}_{rec}+\alpha_{1}\mathcal{L}_{ivc}+\alpha_{2}\mathcal{L}_{pull}+\alpha_{3}\mathcal{L}^{intra}_{push} + \alpha_{4}\mathcal{L}^{inter}_{push} \text{,}
\end{math}
where  $\alpha_{1}$, $\alpha_{2}$, $\alpha_{3}$, and $\alpha_{4}$ are hyper-parameters  to balance losses.

\paragraph{Inference.}
To select the top-1 proposal from the $K$ positive proposals, we use vote-based selection to choose the best overlapping proposal, similar to \cite{zhou2021ensemble, zheng2022cpl}.

\section{Experiments}


\subsection{Experimental Setup}
\label{sec:experimental-setup}

\paragraph{Evaluation metrics.} Following the evaluation metrics in \cite{gao2017tall}, we adopt two metrics (`R@$n$,IoU=$m$' and `R@$n$,mIoU').
`R@$n$,IoU=$m$' denotes the percentage of at least one of the top-$n$ predicted temporal locations having a temporal Intersection over Union (IoU) with a ground truth larger than $m$.
`R@$n$,mIoU' denotes the average of the highest IoUs among the $n$ predicted temporal locations.

\begin{table}[t!]
  \centering
  \resizebox{\columnwidth}{!}{
  \begin{tabular}{l ccc ccc}
    \toprule
    \multirow{2}{*}{Method} & \multicolumn{3}{c}{R@1,IoU=} & \multicolumn{3}{c}{R@5,IoU=} \\ 
     & 0.1 & 0.3 & 0.5 & 0.1 & 0.3 & 0.5\\
    \midrule
    Random & 38.23 & 18.64 & 7.63 & 75.74 & 52.78 & 29.49 \\
    WS-DEC & 62.71 & 41.98 & 23.34 & - & - & - \\
    MARN & - & 47.01 & 29.95 & - & 72.02 & 57.49 \\
    VCA & 67.96 & 50.45 & 31.00 & 92.14 & 71.79 & 53.83 \\
    EC-SL & 68.48 & 44.29 & 24.16 & - & - & - \\
    SCN & 71.48 & 47.23 & 29.22 & 90.88 & 71.56 & 55.69 \\
    RTBPN & 73.73 & 49.77 & 29.63 & 93.89 & 79.89 & 60.56 \\
    LCNet & 78.58 & 48.49 & 26.33 & 93.95 & 82.51 & 62.66 \\
    CCL & - & 50.12 & 31.07 & - & 77.36 & 61.29 \\
    WSTAN & 79.78 & 52.45 & 30.01 & 93.15 & 79.38 & 63.42 \\
    FSAN & 78.45 & 55.11 & 29.43 & 92.59 & 76.79 & 63.32 \\
    CWSTG & 71.86 & 46.62 & 29.52 & 93.75 & 80.92 & 66.61 \\
    CPL & \underbar{82.55} & 55.73 & 31.37 & 87.24 & 63.05 & 43.13 \\
    \midrule
    CRM$^*$ & 81.61 & 55.26 & 32.19 & - & - & - \\
    CNM$^*$ & 78.13 & 55.68 & \underbar{33.33} & - & - & - \\
    IRON$^*$ & \textbf{84.42} & \underbar{58.95} & \textbf{36.27} & \textbf{96.74} & \textbf{85.60} & \underbar{68.52} \\
    \midrule
    PPS & 81.84 & \textbf{59.29} & 31.25 & \underbar{95.28} & \underbar{85.54} & \textbf{71.32} \\
    \bottomrule
  \end{tabular}}
  \caption{Performance comparisons on the ActivityNet Captions. The best results and second best results are represented as bold and underlined numbers, respectively. The methods using additional annotations or large-scale pre-trained models are marked with $^*$.} 
  \label{tab:comparisons-activitynet}
\end{table}

\paragraph{The ActivityNet Captions dataset}~\cite{krishna2017dense}
contains 37,417, 17,505, and 17,031 video-sentence pairs for training, validating $val_1$, and $val_2$, respectively.
Since a testing set is not publicly available, $val_2$ is used for testing.
Video segment features are extracted via C3D~\cite{tran2015learning}.
Vocabulary sizes are 8,000.
For proposals, $K$, $E_{en}$, and $\sigma$ are set to $5$, $2$, and $4$.
For losses, $\alpha_1$, $\alpha_2$, $\alpha_3$, and $\alpha_4$ are set to $1$, $0.2$, $0.01$, and $0.1$.

\paragraph{The Charades-STA dataset}~\cite{gao2017tall}
contains 16,128 video-sentence pairs from 6,672 videos, which are divided into 12,408 for training and 3,720 for testing.
Video segment features are extracted via I3D~\cite{carreira2017quo}.
Vocabulary sizes are 1,111.
For proposals, $K$, $E_{en}$, and $\sigma$ are set to $7$, $3$, and $9$.
For losses, $\alpha_1$, $\alpha_2$, $\alpha_3$, and $\alpha_4$ are set to $3$, $5$, $0.001$, and $1$.

\paragraph{Implementation details.} We set the maximum number of video segments to 200, and the maximum length of the sentence query to 20.
For the transformers, we use transformers with three-layer and four attention heads.
The dimension of the features ($d_V$, $d_Q$, $d_G$, $d_R$) is set to 256.
We use the equivalent MC Transformer for every reconstruction process.
For the hidden sentence query, we randomly hide a third ($1/3$) of the words.
For training, the Adam optimizer~\cite{kingma2014adam} is used.
We set the learning rate to 0.0004, mini-batch size to 32, and hyper-parameters as $\lambda_1=\lambda_2=0.15$, $\beta_1=0.1$, and $\beta_2=0.15$.
In the $k^{th}$ positive proposal, we set the number of Gaussian masks $E_p$ to $k$ for reflecting a varying number of masks in each proposal, as shown in the top right of \cref{fig:framework}.

\begin{table*}[t!]
  \centering
  \resizebox{0.95\linewidth}{!}{
  \begin{tabular}{cc c cccc}
    \toprule
    \multirow{2}{*}{Component} & \multirow{2}{*}{Strategy} & Loss & \multicolumn{2}{c}{R@1} & \multicolumn{2}{c}{R@5} \\ 
     &  & $\mathcal{L}_{pull}$ \& $\mathcal{L}^{intra}_{push}$ & IoU=0.3 & mIoU & IoU=0.3 & mIoU \\
    \midrule
    \multirow{2}{*}{Proposal type} 
     & Single Gaussian & \xmark & 47.49 & 33.33 & 78.23 & 54.85 \\
     & Gaussian mixture & \cmark & \textbf{59.29} & \textbf{37.59} & \textbf{85.54} & \textbf{58.78} \\
    \midrule
    \multirow{3}{*}{Gaussian generation} 
     & Learning one center \& multiple widths & \xmark & 46.63 & 31.54 & 83.65 & \textbf{59.49} \\
     & Learning multiple centers \& widths & \cmark & 47.82 & 32.08 & 84.09 & 56.35 \\ 
     & Learning multiple centers \& one width  & \cmark & \textbf{59.29} & \textbf{37.59} & \textbf{85.54} & 58.78 \\
     \midrule
     \multirow{3}{*}{Importance weighting} & No importance & \cmark & 52.55 & 34.56 & 79.67 & 58.41 \\
     & Importance from the generator & \cmark & 48.55 & 32.36 & 78.96 & 56.87 \\ 
     & Importance from the reconstructor & \cmark & \textbf{59.29} & \textbf{37.59} & \textbf{85.54} & \textbf{58.78} \\ 
    \bottomrule
  \end{tabular}
  }
  \caption{Ablation studies of Gaussian mixture proposals on the ActivityNet Captions dataset.}

  \label{tab:ablation-positive-proposals}
\end{table*}

\begin{table}[t!]
  \centering
  \resizebox{\columnwidth}{!}{
  \begin{tabular}{l ccc ccc}
    \toprule
    \multirow{2}{*}{Method} & \multicolumn{3}{c}{R@1,IoU=} & \multicolumn{3}{c}{R@5,IoU=} \\ 
     & 0.3 & 0.5 & 0.7 & 0.3 & 0.5 & 0.7\\
    \midrule
    Random & 20.12 & 8.61 & 3.39 & 68.42 & 37.57 & 14.98 \\
    TGA & 32.14 & 19.94 & 8.84 & 86.58 & 65.52 & 33.51 \\
    SCN & 42.96 & 23.58 & 9.97 & 95.56 & 71.80 & 38.87 \\
    WSTAN & 43.39 & 29.35 & 12.28 & 93.04 & 76.13 & 41.53 \\
    VLANet & 45.24 & 31.83 & 14.17 & 95.70 & 82.85 & 33.09 \\
    MARN & 48.55 & 31.94 & 14.81 & 90.70 & 70.00 & 37.40 \\
    CCL & - & 33.21 & 15.68 & - & 73.50 & 41.87 \\
    RTBPN & 60.04 & 32.36 & 13.24 & 97.48 & 71.85 & 41.18 \\
    LoGAN & 51.67 & 34.68 & 14.54 & 92.74 & 74.30 & 39.11 \\
    VCA & 58.58 & 38.13 & 19.57 & 98.08 & 78.75 & 37.75 \\
    LCNet & 59.60 & 39.19 & 18.87 & 94.78 & 80.56 & 45.24 \\
    CWSTG & 43.31 & 31.02 & 16.53 & 95.54 & 77.53 & 41.91 \\
    CPL & 66.40 & 49.24 &
    22.39 & 96.99 & 84.71 & 52.37 \\
    \midrule
    CRM$^*$ & 53.66 & 34.76 & 16.37 & -  & - & - \\
    CNM$^*$ & 60.39 & 35.43 & 15.45 & -  & - & - \\
    IRON$^*$ & \textbf{70.71} & \textbf{51.84} & \underbar{25.01} & \underbar{98.96} & \textbf{86.80} & \textbf{54.99} \\
    \midrule
    PPS & \underbar{69.06} & \underbar{51.49} & \textbf{26.16} & \textbf{99.18} & \underbar{86.23} & \underbar{53.01} \\
    \bottomrule
  \end{tabular}}
  \caption{Performance comparisons on the Charades-STA.
  The best results and second best results are represented as bold and underlined numbers, respectively. The methods using additional annotations or large-scale pre-trained models are marked with $^*$.} 
  \label{tab:comparisons-charades}
\end{table}


\subsection{Comparison with State-of-the-Art Methods}
\label{sec:comparison-with-state-of-the-art-methods}
To verify the effectiveness of the proposed method, we compare our PPS with previous weakly supervised temporal video grounding methods:
WS-DEC~\cite{duan2018weakly},
TGA~\cite{mithun2019weakly},
SCN~\cite{lin2020weakly},
WSTAN~\cite{wang2021weakly},
VLANet~\cite{ma2020vlanet},
MARN~\cite{song2020weakly},
CCL~\cite{zhang2020counterfactual},
RTBPN~\cite{zhang2020regularized},
EC-SL~\cite{chen2021towards},
LoGAN~\cite{tan2021logan},
VCA~\cite{wang2021visual},
LCNet~\cite{yang2021local},
FSAN~\cite{wang2022fine},
CWSTG~\cite{Chen_Luo_Zhang_Ma_2022},
CPL~\cite{zheng2022cpl},
CRM~\cite{huang2021cross},
CNM~\cite{zheng2022cnm}, and
IRON~\cite{cao2023iterative}.

In \cref{tab:comparisons-activitynet} for the ActivityNet Captions dataset, our PPS outperforms CPL~\cite{zheng2022cpl} by $3.56\%$, $22.49\%$, and $28.19\%$ at R@1,IoU=0.3, R@5,IoU=0.3, and R@5,IoU=0.5, respectively.
It is worth noting that PPS outperforms the previous learnable mask-based method, CPL, by significant margins at R@5, which means that the generated proposals of PPS promise a higher level of quality.
In \cref{tab:comparisons-charades} for the Charades-STA dataset, our PPS surpasses CPL~\cite{zheng2022cpl} by $3.77\%$ and $2.19\%$ at R@1,IoU=0.7 and R@5,IoU=0.3, respectively.
The methods marked with $^*$ make unfair comparisons with the previous methods.
CRM~\cite{huang2021cross} uses additional paragraph description annotations. CNM~\cite{zheng2022cnm} uses CLIP large-scale pre-trained features~\cite{radford2021learning} and IRON~\cite{cao2023iterative} uses OATrans~\cite{wang2022object} and DistilBERT~\cite{sanh2019distilbert} large-scale pre-trained features.
Although our PPS uses 3D ConvNet and Glove features for fair comparisons with previous methods, PPS shows competitive or higher performance with the methods marked with $^*$.

\begin{figure}[t!]
  \centering
  \includegraphics[width=\linewidth]{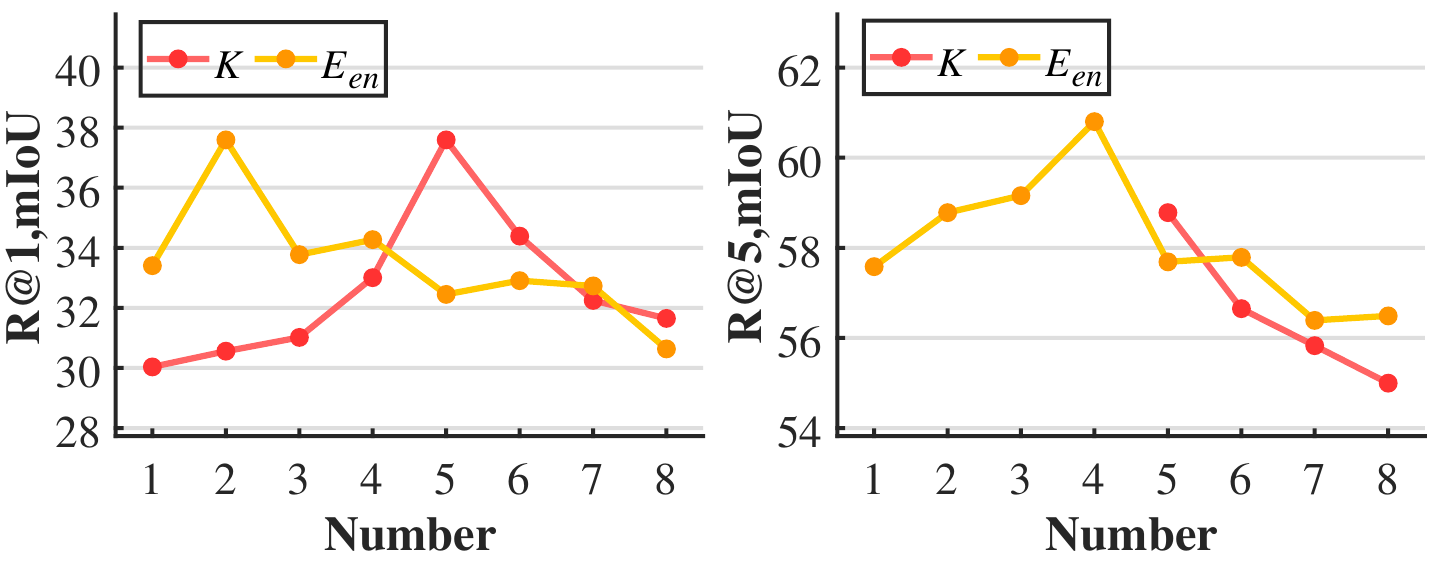}
    \caption{Ablation studies by varying the number of positive and negative proposals $K$ and the number of Gaussian masks of an easy negative proposal $E_{en}$.
    }
    \label{fig:ablation-graph-props}
\end{figure}

\subsection{Ablation Study}
\label{sec:ablation-study}
For a more in-depth understanding of the proposed method, we perform ablation studies on our components.

\paragraph{Analysis on the Gaussian mixture proposal.}
As shown in \cref{tab:ablation-positive-proposals}, we study the impact of the different strategies to generate Gaussian mixture proposals for positive proposals.
The results are summarized as follows:
First, the Gaussian mixture proposals are more effective than the single Gaussian proposal, which means that the mixture proposal can better represent a query-relevant temporal location.
Second, learning multiple centers and one width for one mixture proposal performs best.
We conjecture that learning multiple widths makes it complicated to learn proposals, which reduces performance.
Third, importance weighting from the reconstructor yields the best result by representing the importance of each mask for query reconstruction.
On the other hand, importance weighting from the generator is less effective, because it is hard to reflect reconstruction-aware information.
\cref{fig:ablation-graph-props} shows the impact of the number of proposals $K$.
The performance increases until the number is $5$ at R@1,mIoU.
We observe that defining too many proposals makes the proposals redundant and have short lengths due to the impact of the inter-pushing loss $\mathcal{L}^{inter}_{push}$.

\paragraph{Impact on a varying number of masks.}
For positive proposals, we form $\mathbf{P}_{p}^{(k)}$ by a Gaussian mixture of $E_p=k$ Gaussian masks to reflect a varying number of Gaussian masks in each positive proposal.
To verify the effectiveness of the varying number of Gaussian masks, we compare the performance of fixing the number of Gaussian masks for every positive proposal in \cref{tab:ablation-positive-proposals-fixing}.
The results show that using a varying number of Gaussian masks for each positive proposal performs better than using a fixed Gaussian number of Gaussian masks.
We find that combinations of different numbers of Gaussian masks can represent a diverse number of query-relevant events.

\begin{table}[t!]
  \centering
          \begin{subtable}[t]{.47\linewidth}
            \centering
              \resizebox{\columnwidth}{!}{
              \begin{tabular}{c cc}
                \toprule
                \multirow{2}{*}{\# masks} & R@1 & R@5 \\ 
                 & mIoU & mIoU \\
                \midrule
                Fix to 1 & 33.33 & 54.85 \\ 
                Fix to 3 & 35.91 & 56.27 \\ 
                Fix to 5 & 36.58 & 54.36 \\ 
                Fix to 7 & 36.25 & 49.48 \\ 
                \midrule
                Vary & \textbf{37.59} & \textbf{58.78} \\ 
                \bottomrule
          \end{tabular}}
        \caption{The number of masks for a positive proposal.}
          \label{tab:ablation-positive-proposals-fixing}
            \end{subtable}%
        \hspace{0.2cm}
        \begin{subtable}[t]{.47\linewidth}
                  \centering
                      \resizebox{\columnwidth}{!}{
                      \begin{tabular}{c cc}
                        \toprule
                        Pulling & R@1 & R@5 \\ 
                        strategy & mIoU & mIoU \\
                        \midrule
                        All & 35.41 & 56.87 \\ 
                        To mid & 35.83 & \textbf{58.92} \\ 
                        Distant & \textbf{37.59} & 58.78 \\ 
                        \bottomrule
                \end{tabular}}
                \caption{Strategies for the pulling loss}
              \label{tab:ablation-pulling-loss}
        \end{subtable}%
        \caption{Ablation studies on the ActivityNet Captions dataset.}
          
  \label{tab:ablation-others}
\end{table}

\begin{table}[t!]
  \centering
  \resizebox{\columnwidth}{!}{
  \begin{tabular}{ccc cccc}
    \toprule
    \multicolumn{3}{c}{Loss} & \multicolumn{2}{c}{R@1,IoU=} & \multicolumn{2}{c}{R@5,IoU=} \\ 
    $\mathcal{L}_{pull}$ & $\mathcal{L}^{intra}_{push}$ & $\mathcal{L}^{inter}_{push}$ & 0.3 & mIoU & 0.3 & mIoU \\
    \midrule
    \xmark & \xmark & \xmark & 45.23 & 30.98 & 67.86 & 47.75 \\
    \cmark & \xmark & \xmark & 55.03 & 36.79 & 72.49 & 51.14 \\
    \xmark & \cmark & \xmark & 45.69 & 29.83 & 71.51 & 49.69 \\
    \xmark & \xmark & \cmark & 23.50 & 15.96 & 72.30 & 44.46 \\
    \xmark & \cmark & \cmark & 23.47 & 16.38 & 66.15 & 38.20 \\
    \cmark & \xmark & \cmark & 41.51 & 30.23 & 85.18 & \textbf{60.44} \\
    \cmark & \cmark & \xmark & 49.98 & 32.46 & 80.54 & 55.26 \\ 
    \cmark & \cmark & \cmark & \textbf{59.29} & \textbf{37.59} & \textbf{85.54} & 58.78 \\ 
    \bottomrule
  \end{tabular}}
  \caption{Ablation studies of different losses for the pull-push learning scheme on the ActivityNet Captions dataset.}
  \label{tab:ablation-loss}
\end{table}

\begin{figure}[t!]
  \centering
    \includegraphics[width=\linewidth]{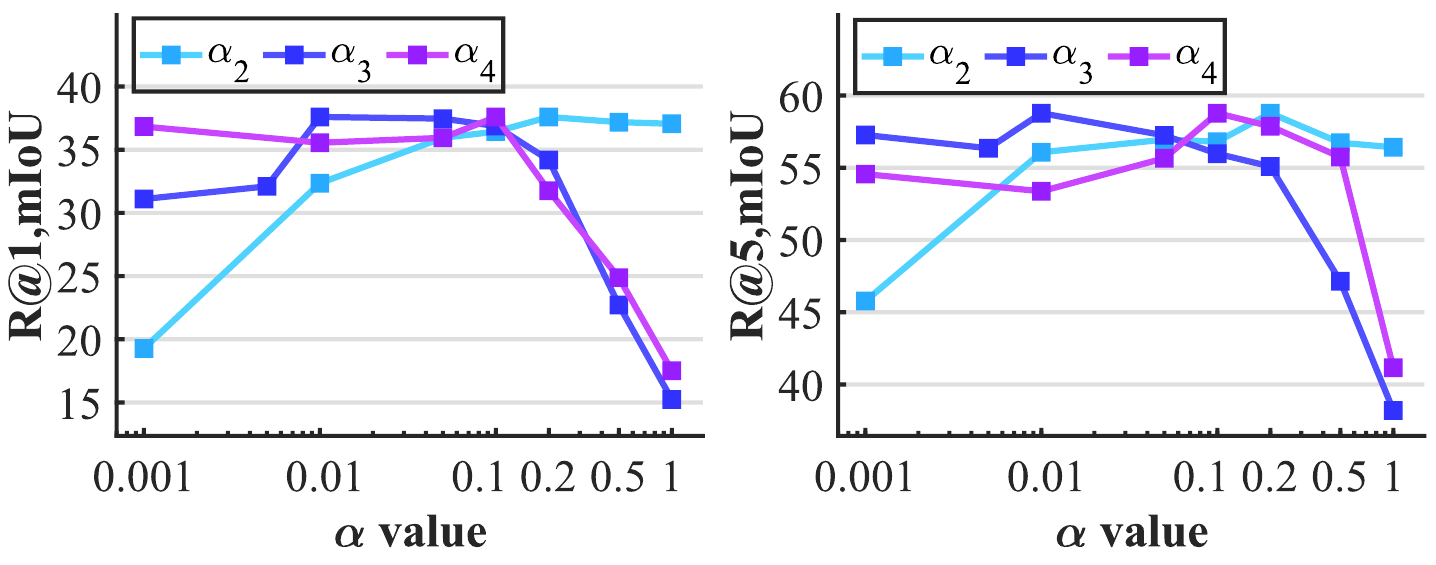}
    \caption{Ablation studies by varying $\alpha$ values for the pull-push learning scheme on the ActivityNet Captions dataset.
    }
    \label{fig:ablation-graph-alpha}
\end{figure}

\paragraph{Effect of the pull-push learning scheme.}
In \cref{tab:ablation-loss}, we verify the effectiveness of our pull-push learning scheme.
Among combinations of three losses ($\mathcal{L}_{pull}$, $\mathcal{L}^{intra}_{push}$, $\mathcal{L}^{inter}_{push}$), adopting all three losses yields the best performance.
We conjecture that our pull-push learning scheme helps Gaussian masks to capture diverse events for better representing a temporal location.
It is notable that adopting only the pulling loss can yield competitive or higher results to the state-of-the-art methods in \cref{tab:comparisons-activitynet}.
If the pulling loss $\mathcal{L}_{pull}$ is excluded, the performance decreases significantly.
We observe that Gaussian masks for one Gaussian mixture proposal are spread sparsely throughout the entire video without $\mathcal{L}_{pull}$, which can not represent one proper temporal location.
Additionally, the results suggest that two pushing losses ($\mathcal{L}^{intra}_{push}$, $\mathcal{L}^{inter}_{push}$) are used with $\mathcal{L}_{pull}$ for a synergy effect, because the goal of the pushing losses is to make less overlapped masks for moderate coupling.
For a more in-depth understanding of the pulling loss $\mathcal{L}_{pull}$, we conduct ablation studies of different strategies for $\mathcal{L}_{pull}$ in \cref{tab:ablation-pulling-loss}.
Among the strategies, pulling two distant masks closer or pulling two distant masks to the middle mask performs best.
The results imply that pulling fewer masks is better and pulling more masks may ruin the structure of the mixture proposal due to overlapped masks.
\cref{fig:ablation-graph-alpha} presents the impact of controlling the balance of the losses.
The results show that a high $\alpha_2$ value for $\mathcal{L}_{pull}$ is needed to produce densely generated masks and the adequate $\alpha_3$ and $\alpha_4$ values for $\mathcal{L}^{intra}_{push}$ and $\mathcal{L}^{inter}_{push}$ are needed to cause proper discrimination between the masks and between the proposals, respectively.

\subsection{Qualitative Results}
\label{sec:qualitative-results}

\begin{figure}[t]
  \centering
  \includegraphics[width=\linewidth]{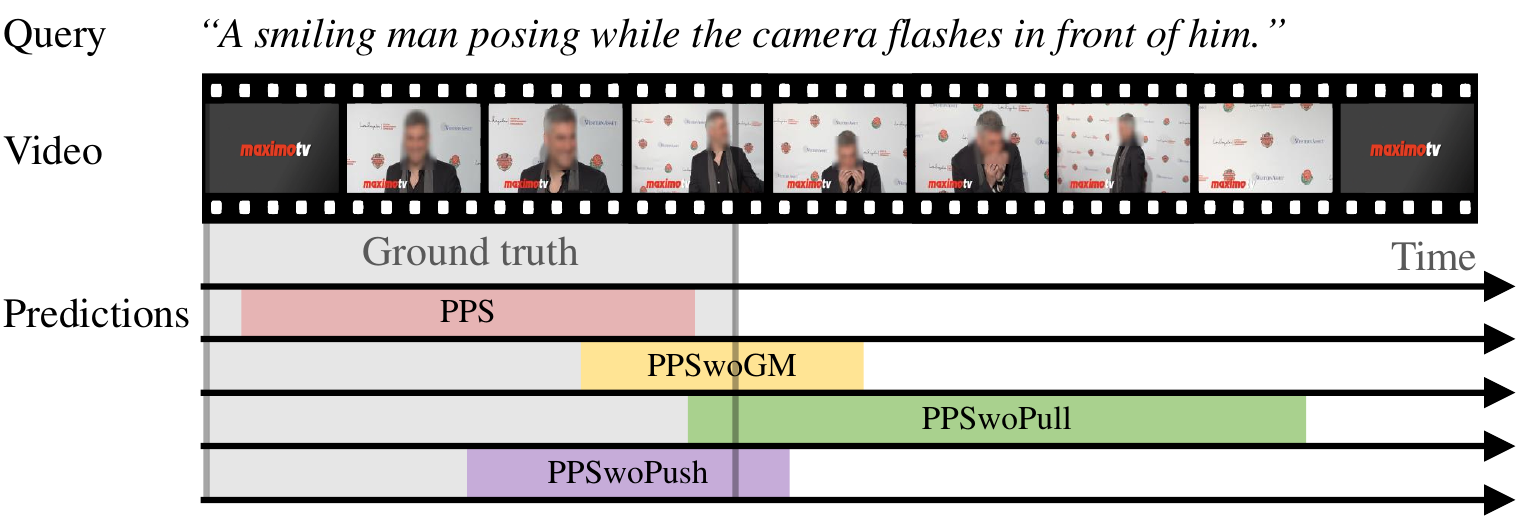}
  \caption{
  Qualitative results on the Activity-Net Captions dataset.
  Given a video and a query, PPS yields a predicted temporal location (red).
  We also visualize the predictions of variants using a positive proposal of one Gaussian mask without the mixture (yellow) or excluding a pulling loss (green) or excluding pushing losses (purple).
  }
\label{fig:qualitative}
\end{figure}

\cref{fig:qualitative} shows qualitative results of our PPS and other variants of PPS.
It is notable that PPS captures accurate query-relevant locations, while the ground truth, which can be noisy due to the subjectivity of annotators, includes redundant locations such as a logo at the beginning of the video.

\section{Conclusion}
For weakly supervised temporal video grounding, we have proposed Gaussian mixture proposals with a pull-push learning scheme to capture diverse events.
We express arbitrary shapes of a temporal location by learning importance, centroid, and range of every Gaussian in the mixture.
To produce moderately coupled Gaussians in the mixture, we leverage a pulling loss and a pushing loss, each of which plays an opposite role to the other.
Through experimental comparisons and extensive ablation studies, we have verified that our method generates multiple high-quality proposals, which greatly improve recall rates.

\paragraph{Limitations.}
We use the proposals with the shape of a Gaussian mixture, but other shapes could be explored to represent complex temporal structures.

\section*{Acknowledgments}
This work was supported by IITP grant funded by Korea government(MSIT) [No.B0101-15-0266, Development of High Performance Visual BigData Discovery Platform for Large-Scale Realtime Data Analysis; NO.2021-0-01343, Artificial Intelligence Graduate School Program (Seoul National University)] and the BK21 FOUR program of the Education and Research Program for Future ICT Pioneers, Seoul National University in 2023.


\bibliography{aaai24}

\begin{thebibliography}{44}
\providecommand{\natexlab}[1]{#1}

\bibitem[{Cao et~al.(2023)Cao, Wei, Xu, Geng, Chen, Zhang, Zou, Shen, and
  Jiang}]{cao2023iterative}
Cao, M.; Wei, F.; Xu, C.; Geng, X.; Chen, L.; Zhang, C.; Zou, Y.; Shen, T.; and
  Jiang, D. 2023.
\newblock Iterative Proposal Refinement for Weakly-Supervised Video Grounding.
\newblock In \emph{CVPR}, 6524--6534.

\bibitem[{Carreira and Zisserman(2017)}]{carreira2017quo}
Carreira, J.; and Zisserman, A. 2017.
\newblock Quo vadis, action recognition? a new model and the kinetics dataset.
\newblock In \emph{CVPR}, 6299--6308.

\bibitem[{Chen et~al.(2022)Chen, Luo, Zhang, and Ma}]{Chen_Luo_Zhang_Ma_2022}
Chen, J.; Luo, W.; Zhang, W.; and Ma, L. 2022.
\newblock Explore Inter-contrast between Videos via Composition for Weakly
  Supervised Temporal Sentence Grounding.
\newblock In \emph{AAAI}.

\bibitem[{Chen and Jiang(2021)}]{chen2021towards}
Chen, S.; and Jiang, Y.-G. 2021.
\newblock Towards bridging event captioner and sentence localizer for weakly
  supervised dense event captioning.
\newblock In \emph{CVPR}, 8425--8435.

\bibitem[{Devlin et~al.(2019)Devlin, Chang, Lee, and
  Toutanova}]{devlin2018bert}
Devlin, J.; Chang, M.-W.; Lee, K.; and Toutanova, K. 2019.
\newblock Bert: Pre-training of deep bidirectional transformers for language
  understanding.
\newblock In \emph{NAACL-HLT}.

\bibitem[{Dong et~al.(2019)Dong, Li, Xu, Ji, He, Yang, and Wang}]{dong2019dual}
Dong, J.; Li, X.; Xu, C.; Ji, S.; He, Y.; Yang, G.; and Wang, X. 2019.
\newblock Dual encoding for zero-example video retrieval.
\newblock In \emph{CVPR}, 9346--9355.

\bibitem[{Duan et~al.(2018)Duan, Huang, Gan, Wang, Zhu, and
  Huang}]{duan2018weakly}
Duan, X.; Huang, W.; Gan, C.; Wang, J.; Zhu, W.; and Huang, J. 2018.
\newblock Weakly supervised dense event captioning in videos.
\newblock In \emph{NeurIPS}, volume~31.

\bibitem[{Gao et~al.(2017)Gao, Sun, Yang, and Nevatia}]{gao2017tall}
Gao, J.; Sun, C.; Yang, Z.; and Nevatia, R. 2017.
\newblock Tall: Temporal activity localization via language query.
\newblock In \emph{ICCV}, 5267--5275.

\bibitem[{Huang et~al.(2021)Huang, Liu, Gong, and Jin}]{huang2021cross}
Huang, J.; Liu, Y.; Gong, S.; and Jin, H. 2021.
\newblock Cross-sentence temporal and semantic relations in video activity
  localisation.
\newblock In \emph{ICCV}, 7199--7208.

\bibitem[{Kim et~al.(2022)Kim, Ha, Yun, and Choi}]{kim2022swag}
Kim, S.; Ha, T.; Yun, K.; and Choi, J.~Y. 2022.
\newblock SWAG-Net: Semantic Word-Aware Graph Network for Temporal Video
  Grounding.
\newblock In \emph{ACM CIKM}, 982–992.

\bibitem[{Kim, Yun, and Choi(2021)}]{kim2021plrn}
Kim, S.; Yun, K.; and Choi, J.~Y. 2021.
\newblock Position-aware Location Regression Network for Temporal Video
  Grounding.
\newblock In \emph{AVSS}, 1--8.

\bibitem[{Kingma and Ba(2015)}]{kingma2014adam}
Kingma, D.~P.; and Ba, J. 2015.
\newblock Adam: A method for stochastic optimization.
\newblock In \emph{ICLR}.

\bibitem[{Krishna et~al.(2017)Krishna, Hata, Ren, Fei-Fei, and
  Carlos~Niebles}]{krishna2017dense}
Krishna, R.; Hata, K.; Ren, F.; Fei-Fei, L.; and Carlos~Niebles, J. 2017.
\newblock Dense-captioning events in videos.
\newblock In \emph{ICCV}, 706--715.

\bibitem[{Lee et~al.(2018)Lee, Lee, Lee, and Shin}]{lee2018simple}
Lee, K.; Lee, K.; Lee, H.; and Shin, J. 2018.
\newblock A simple unified framework for detecting out-of-distribution samples
  and adversarial attacks.
\newblock \emph{NeurIPS}, 31.

\bibitem[{Lin et~al.(2017)Lin, Feng, Santos, Yu, Xiang, Zhou, and
  Bengio}]{lin2017structured}
Lin, Z.; Feng, M.; Santos, C. N.~d.; Yu, M.; Xiang, B.; Zhou, B.; and Bengio,
  Y. 2017.
\newblock A structured self-attentive sentence embedding.
\newblock \emph{arXiv preprint arXiv:1703.03130}.

\bibitem[{Lin et~al.(2020)Lin, Zhao, Zhang, Wang, and Liu}]{lin2020weakly}
Lin, Z.; Zhao, Z.; Zhang, Z.; Wang, Q.; and Liu, H. 2020.
\newblock Weakly-supervised video moment retrieval via semantic completion
  network.
\newblock In \emph{AAAI}, volume~34, 11539--11546.

\bibitem[{Long et~al.(2019)Long, Yao, Qiu, Tian, Luo, and
  Mei}]{long2019gaussian}
Long, F.; Yao, T.; Qiu, Z.; Tian, X.; Luo, J.; and Mei, T. 2019.
\newblock Gaussian temporal awareness networks for action localization.
\newblock In \emph{CVPR}, 344--353.

\bibitem[{Ma et~al.(2020)Ma, Yoon, Kim, Lee, Kang, and Yoo}]{ma2020vlanet}
Ma, M.; Yoon, S.; Kim, J.; Lee, Y.; Kang, S.; and Yoo, C.~D. 2020.
\newblock Vlanet: Video-language alignment network for weakly-supervised video
  moment retrieval.
\newblock In \emph{ECCV}, 156--171. Springer.

\bibitem[{Ma et~al.(2002)Ma, Lu, Zhang, and Li}]{ma2002user}
Ma, Y.-F.; Lu, L.; Zhang, H.-J.; and Li, M. 2002.
\newblock A user attention model for video summarization.
\newblock In \emph{ACM MM}, 533--542.

\bibitem[{Mithun, Paul, and Roy-Chowdhury(2019)}]{mithun2019weakly}
Mithun, N.~C.; Paul, S.; and Roy-Chowdhury, A.~K. 2019.
\newblock Weakly supervised video moment retrieval from text queries.
\newblock In \emph{CVPR}, 11592--11601.

\bibitem[{Neubeck and Van~Gool(2006)}]{neubeck2006efficient}
Neubeck, A.; and Van~Gool, L. 2006.
\newblock Efficient non-maximum suppression.
\newblock In \emph{ICPR}, volume~3, 850--855. IEEE.

\bibitem[{Pennington, Socher, and Manning(2014)}]{pennington2014glove}
Pennington, J.; Socher, R.; and Manning, C. 2014.
\newblock {G}lo{V}e: Global Vectors for Word Representation.
\newblock In \emph{EMNLP}, 1532--1543.

\bibitem[{Piergiovanni and Ryoo(2019)}]{piergiovanni2019temporal}
Piergiovanni, A.; and Ryoo, M. 2019.
\newblock Temporal gaussian mixture layer for videos.
\newblock In \emph{ICML}, 5152--5161. PMLR.

\bibitem[{Radford et~al.(2021)Radford, Kim, Hallacy, Ramesh, Goh, Agarwal,
  Sastry, Askell, Mishkin, Clark et~al.}]{radford2021learning}
Radford, A.; Kim, J.~W.; Hallacy, C.; Ramesh, A.; Goh, G.; Agarwal, S.; Sastry,
  G.; Askell, A.; Mishkin, P.; Clark, J.; et~al. 2021.
\newblock Learning transferable visual models from natural language
  supervision.
\newblock In \emph{ICML}, 8748--8763. PMLR.

\bibitem[{Sanh et~al.(2019)Sanh, Debut, Chaumond, and
  Wolf}]{sanh2019distilbert}
Sanh, V.; Debut, L.; Chaumond, J.; and Wolf, T. 2019.
\newblock DistilBERT, a distilled version of BERT: smaller, faster, cheaper and
  lighter.
\newblock \emph{arXiv preprint arXiv:1910.01108}.

\bibitem[{Song et~al.(2020)Song, Wang, Ma, Yu, and Yu}]{song2020weakly}
Song, Y.; Wang, J.; Ma, L.; Yu, Z.; and Yu, J. 2020.
\newblock Weakly-supervised multi-level attentional reconstruction network for
  grounding textual queries in videos.
\newblock \emph{arXiv preprint arXiv:2003.07048}.

\bibitem[{Tan et~al.(2021)Tan, Xu, Saenko, and Plummer}]{tan2021logan}
Tan, R.; Xu, H.; Saenko, K.; and Plummer, B.~A. 2021.
\newblock Logan: Latent graph co-attention network for weakly-supervised video
  moment retrieval.
\newblock In \emph{WACV}, 2083--2092.

\bibitem[{Tran et~al.(2015)Tran, Bourdev, Fergus, Torresani, and
  Paluri}]{tran2015learning}
Tran, D.; Bourdev, L.; Fergus, R.; Torresani, L.; and Paluri, M. 2015.
\newblock Learning spatiotemporal features with 3d convolutional networks.
\newblock In \emph{ICCV}, 4489--4497.

\bibitem[{Vaswani et~al.(2017)Vaswani, Shazeer, Parmar, Uszkoreit, Jones,
  Gomez, Kaiser, and Polosukhin}]{vaswani2017attention}
Vaswani, A.; Shazeer, N.; Parmar, N.; Uszkoreit, J.; Jones, L.; Gomez, A.~N.;
  Kaiser, {\L}.; and Polosukhin, I. 2017.
\newblock Attention is all you need.
\newblock In \emph{NeurIPS}, volume~30.

\bibitem[{Wang et~al.(2022)Wang, Ge, Cai, Yan, Lin, Shan, Qie, and
  Shou}]{wang2022object}
Wang, J.; Ge, Y.; Cai, G.; Yan, R.; Lin, X.; Shan, Y.; Qie, X.; and Shou, M.~Z.
  2022.
\newblock Object-aware video-language pre-training for retrieval.
\newblock In \emph{CVPR}, 3313--3322.

\bibitem[{Wang et~al.(2014)Wang, Song, Leung, Rosenberg, Wang, Philbin, Chen,
  and Wu}]{wang2014learning}
Wang, J.; Song, Y.; Leung, T.; Rosenberg, C.; Wang, J.; Philbin, J.; Chen, B.;
  and Wu, Y. 2014.
\newblock Learning fine-grained image similarity with deep ranking.
\newblock In \emph{CVPR}, 1386--1393.

\bibitem[{Wang et~al.(2021)Wang, Deng, Zhou, and Li}]{wang2021weakly}
Wang, Y.; Deng, J.; Zhou, W.; and Li, H. 2021.
\newblock Weakly supervised temporal adjacent network for language grounding.
\newblock \emph{IEEE Transactions on Multimedia}.

\bibitem[{Wang, Zhou, and Li(2021)}]{wang2022fine}
Wang, Y.; Zhou, W.; and Li, H. 2021.
\newblock Fine-grained semantic alignment network for weakly supervised
  temporal language grounding.
\newblock In \emph{EMNLP}.

\bibitem[{Wang, Chen, and Jiang(2021)}]{wang2021visual}
Wang, Z.; Chen, J.; and Jiang, Y.-G. 2021.
\newblock Visual co-occurrence alignment learning for weakly-supervised video
  moment retrieval.
\newblock In \emph{ACM MM}, 1459--1468.

\bibitem[{Wu et~al.(2020)Wu, Li, Han, and Lin}]{wu2020reinforcement}
Wu, J.; Li, G.; Han, X.; and Lin, L. 2020.
\newblock Reinforcement learning for weakly supervised temporal grounding of
  natural language in untrimmed videos.
\newblock In \emph{ACM MM}, 1283--1291.

\bibitem[{Yang et~al.(2021)Yang, Zhang, Zhang, and Wu}]{yang2021local}
Yang, W.; Zhang, T.; Zhang, Y.; and Wu, F. 2021.
\newblock Local correspondence network for weakly supervised temporal sentence
  grounding.
\newblock \emph{IEEE Transactions on Image Processing}, 30: 3252--3262.

\bibitem[{Yuan et~al.(2021)Yuan, Lan, Wang, Chen, Wang, and
  Zhu}]{yuan2021closer}
Yuan, Y.; Lan, X.; Wang, X.; Chen, L.; Wang, Z.; and Zhu, W. 2021.
\newblock A closer look at temporal sentence grounding in videos: Dataset and
  metric.
\newblock In \emph{Proceedings of the 2nd International Workshop on
  Human-centric Multimedia Analysis}, 13--21.

\bibitem[{Zhang et~al.(2020{\natexlab{a}})Zhang, Lin, Zhao, Zhu, and
  He}]{zhang2020regularized}
Zhang, Z.; Lin, Z.; Zhao, Z.; Zhu, J.; and He, X. 2020{\natexlab{a}}.
\newblock Regularized two-branch proposal networks for weakly-supervised moment
  retrieval in videos.
\newblock In \emph{ACM MM}, 4098--4106.

\bibitem[{Zhang et~al.(2020{\natexlab{b}})Zhang, Zhao, Lin, He
  et~al.}]{zhang2020counterfactual}
Zhang, Z.; Zhao, Z.; Lin, Z.; He, X.; et~al. 2020{\natexlab{b}}.
\newblock Counterfactual contrastive learning for weakly-supervised
  vision-language grounding.
\newblock In \emph{NeurIPS}, volume~33, 18123--18134.

\bibitem[{Zheng et~al.(2022{\natexlab{a}})Zheng, Huang, Chen, and
  Liu}]{zheng2022cnm}
Zheng, M.; Huang, Y.; Chen, Q.; and Liu, Y. 2022{\natexlab{a}}.
\newblock Weakly supervised video moment localization with contrastive negative
  sample mining.
\newblock In \emph{AAAI}, volume~1, 3.

\bibitem[{Zheng et~al.(2022{\natexlab{b}})Zheng, Huang, Chen, Peng, and
  Liu}]{zheng2022cpl}
Zheng, M.; Huang, Y.; Chen, Q.; Peng, Y.; and Liu, Y. 2022{\natexlab{b}}.
\newblock Weakly Supervised Temporal Sentence Grounding With Gaussian-Based
  Contrastive Proposal Learning.
\newblock In \emph{CVPR}, 15555--15564.

\bibitem[{Zhou et~al.(2021)Zhou, Zhang, Luo, Chen, and Hu}]{zhou2021embracing}
Zhou, H.; Zhang, C.; Luo, Y.; Chen, Y.; and Hu, C. 2021.
\newblock Embracing uncertainty: Decoupling and de-bias for robust temporal
  grounding.
\newblock In \emph{CVPR}, 8445--8454.

\bibitem[{Zhou(2021)}]{zhou2021ensemble}
Zhou, Z.-H. 2021.
\newblock Ensemble learning.
\newblock In \emph{Machine learning}, 181--210. Springer.

\bibitem[{Zong et~al.(2018)Zong, Song, Min, Cheng, Lumezanu, Cho, and
  Chen}]{zong2018deep}
Zong, B.; Song, Q.; Min, M.~R.; Cheng, W.; Lumezanu, C.; Cho, D.; and Chen, H.
  2018.
\newblock Deep autoencoding gaussian mixture model for unsupervised anomaly
  detection.
\newblock In \emph{ICLR}.

\end{thebibliography}


\end{document}